\documentclass[12pt]{article}

\usepackage{sbc-template}
\usepackage{graphicx,url}
\usepackage[utf8]{inputenc}

\usepackage[table,xcdraw]{xcolor}
\usepackage{subfig}

\title{Enhanced Self-Organizing Map Solution \\for the Traveling Salesman Problem}

\author{Joao~P.~A.~Dantas\inst{1}, Andre~N.~Costa\inst{1}, Marcos~R.~O.~A.~Maximo\inst{2}, Takashi~Yoneyama\inst{3}}

\address{Decision Support Systems Subdivision, Institute for Advanced Studies\\
12.288-001 -- Sao Jose dos Campos -- SP -- Brazil
\nextinstitute
Autonomous Computational System Lab, Aeronautics Institute of Technology\\
12.228-900 -- Sao Jose dos Campos -- SP -- Brazil
\nextinstitute
Electronic Engineering Division, Aeronautics Institute of Technology\\
12.228-900 -- Sao Jose dos Campos -- SP -- Brazil
\email{\{dantasjpad,negraoanc\}@fab.mil.br, \{mmaximo,takashi\}@ita.br}
}

\begin{document} 

\maketitle

\begin{abstract}
Using an enhanced Self-Organizing Map method, we provided suboptimal solutions to the Traveling Salesman Problem. Besides, we employed hyperparameter tuning to identify the most critical features in the algorithm. All improvements in the benchmark work brought consistent results and may inspire future efforts to improve this algorithm and apply it to different problems.
\end{abstract}

\section{Introduction}


The Traveling Salesman Problem (TSP) is one of the most studied routing problems within the combinatorial optimization field~\cite{laporte1992traveling}, which is defined as a salesman that must visit a series of cities once and only once, returning to the starting city~\cite{brocki2007kohonen}. Although coming from a relatively simple idea, finding the solution for the problem, i.e., the shortest path, is NP-Hard~\cite{huang2017improved}, thence the great interest to find efficient ways to solve it.



Due to the problem's complexity and to the fact that many of its applications require fast ways to solve it, it is common to employ heuristics to generate approximate (suboptimal) solutions to the TSP. Examples of these methods are Genetic Algorithm (GA), Simulated Annealing (SA), tabu search, and ant colony optimization~\cite{xu2008self, das2011problema}. Another form to find approximate solutions to complex optimization problems is to use neural networks~\cite{peterson1990parallel}, which can be efficient due to their adaptability.


According to the literature, the first neural network solution for the TSP was the one presented in~\cite{hopfield1985neural}, based on the minimization of an energy function. From that, many methods have been proposed to improve the Hopfield neural network~\cite{xu2008self}. Another use of neural networks for solving the TSP is the one proposed in~\cite{kohonen1998self} through competitive unsupervised learning based on winner-take-all and winner-take-most algorithms, also called Self-Organizing Map (SOM), due to how the adaptation of the neurons works~\cite{brocki2007kohonen}. The SOM adjusts its neurons to fit the input cooperatively by inspecting the list of cities. This localized response to the input list generates a neighborhood preserving map, resulting in a near-optimal path~\cite{xu2008self}. 

In summary, most of the TSP heuristics solutions in the literature have used one of the following methods: Hopfield network, Kohonen’s SOM, GA, and SA~\cite{markovic2012solving}. In comparison with other heuristic methods, the SOM has presented low computation complexity and promising performance. Therefore, these neural networks have attracted researchers to explore and enhance their performance when applied to the TSP~\cite{favata1991study, budinich1996self}.


In our work, the main contribution is to modify the method of SOM as proposed and implemented in~\cite{martin2018}, adding some features and tuning hyperparameters. We offer these modifications to solve a particular problem involving TSPs varying in size from $50$ to $200$. The remainder of this paper is structured as follows: Section~\ref{exp} provides details with respect to the experiment setting; Section~\ref{imp} states the improvements on the method for this context; Section~\ref{res} presents the results obtained; Section~\ref{conc} brings the conclusions and the possible future work.

\section{Experiment}\label{exp}

The proposed experiment aimed at predicting edges belonging to optimal solutions of $2{,}000$ two-dimensional Euclidean TSP ranging from $50$ to $200$ city nodes. We evaluate the solution considering the F1 score of the predicted adjacency matrix compared to the optimal solution. The F1 score is the harmonic mean of the precision and recall, where an F1 score reaches its best value at $1$ and worst score at $0$~\cite{zhang2015estimating}. The relative contribution of precision and recall to the F1 score is equal, which is preferred over accuracy as this problem is often very imbalanced~\cite{mele2021new}.

\section{Improvements}\label{imp}

In this section, we discuss the main improvements performed in the method adopted. We employ hyperparameter tuning to adapt the algorithm. Besides, we created two additional features to find solutions with better results.

\subsection{Hyperparameter Tuning}

When exploring the algorithm used in this work, we first identified its primary hyperparameters (population size, number of iterations, learning rate, and discount rates for the latter two) and their effects on the final scores. To understand the influence of each of these factors and find their best configuration, we employed a single-factor design to tune the proposed technique, i.e., varying approximately one feature at a time to examine its isolated influence. The search for the hyperparameters that better suit the dataset led us to the baseline algorithm, which received all modifications discussed next.

\subsection{Additional Modifications}
\label{additional-features}

The first improvement to the baseline algorithm was to change the way to choose the first node considered. The standard technique randomly selects the starting point, which may lead the solution into a local minimum, depending on the location of this first node. In addition to the randomly chosen initial city, we forced the algorithm to start from the city in the centermost position and the furthest position from the centroid of all cities. After running the algorithm in these three different initial positions, we chose the one that presents the shortest path.

\begin{figure}[hhh]
\centering
\includegraphics[width=\textwidth]{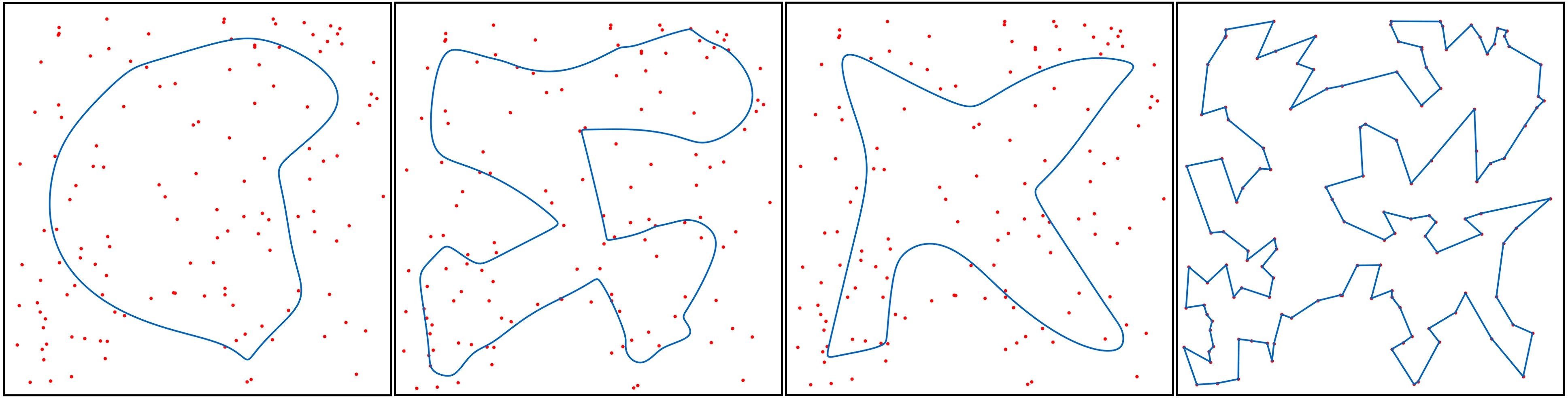}
\caption{SOM algorithm iteration samples.}
\label{fig:Fig1}
\end{figure}

As the second modification, after employing the hyperparameters tuning and identifying the most significant feature as the population size, we improved the algorithm based on the variation of this hyperparameter in each SOM iteration. Figure~\ref{fig:Fig1} exemplifies the SOM iterations until the result with the shortest path is obtained. We used 20 different population sizes in each TSP, from 1 to 20 times the number of cities, and we calculated the route length of all of these TSP solutions, choosing the one with the shortest path. Indeed, that additional feature turned the process computationally more costly. However, it brought better results since the population size was constantly changing, and we chose the one that produced the shortest distance in all solutions.

\section{Results}\label{res}

Table~\ref{tab:table1} shows the F1 Scores results of the hyperparameters tuning and clarifies the importance of the population size feature. We chose as our baseline the following configuration: $100{,}000$ iterations, $0.9997$ for the discount rate of initial neighborhood, $0.8$ for the learning rate, $0.99997$ for the discount rate of the learning rate, and $6$ for the population size multiplier factor.


\begin{table}[h]
\centering
\caption{Hyperparameters tuning with F1 scores}
\label{tab:table1}
\resizebox{0.9\textwidth}{!}{
\begin{tabular}{|c|c|c|c|c|c|}
\hline
\textbf{Number of} & \textbf{Discount Rate of} & \textbf{Learning} & \textbf{Discount Rate of} & \textbf{Population Size} & \textbf{F1} \\
\textbf{Iterations} & \textbf{Initial Neighborhood}  & \textbf{Rate} & \textbf{Learning Rate} & \textbf{Multiplier Factor} &  \textbf{Score}
\\ \hline
100    & 0.9997  & 0.8  & 0.99997 & 8  & \cellcolor[HTML]{E67C73}0.06878 \\ \hline
100{,}000 & 0.9997  & 0.8  & 0.99997 & 8  & \cellcolor[HTML]{FDD167}0.07800 \\ \hline
100{,}000  & 0.9997  & 0.8  & 0.99997 & 16 & \cellcolor[HTML]{FDCF67}0.07784 \\ \hline
100{,}000  & 0.9997  & 0.8  & 0.99997 & 4  & \cellcolor[HTML]{F5D568}0.07855 \\ \hline
100{,}000  & 0.9997  & 0.8  & 0.99997 & 2  & \cellcolor[HTML]{FFD666}0.07853 \\ \hline
100{,}000  & 0.9997  & 0.8  & 0.99997 & 6  & \cellcolor[HTML]{57BB8A}0.07885 \\ \hline
100{,}000  & 0.9997  & 0.8  & 0.99997 & 5  & \cellcolor[HTML]{ACC977}0.07869 \\ \hline
100{,}000  & 0.99997 & 0.8  & 0.99997 & 6  & \cellcolor[HTML]{FAD667}0.07854 \\ \hline
100{,}000  & 0.997   & 0.8  & 0.99997 & 6  & \cellcolor[HTML]{FED467}0.07834 \\ \hline
100{,}000  & 0.997   & 0.01 & 1.00000 & 6  & \cellcolor[HTML]{FCCE68}0.07769 \\ \hline
100{,}000  & 0.9997  & 0.8  & 0.99997 & 7  & \cellcolor[HTML]{7CC182}0.07878 \\ \hline
\end{tabular}
}
\end{table}

After choosing the baseline hyperparameter configuration, we employed the additional modifications described in subsection~\ref{additional-features}, achieving slightly better results with $0.07891$ of the evaluation metric proposed. Therefore, these results show that, through subtle and well-planned changes in algorithms already consolidated by the scientific community, it is possible to obtain better results in classical optimization problems.

\section{Conclusions}\label{conc}

This work brings an enhanced way to use SOM to find suboptimal solutions for the TSP. Firstly, we employ hyperparameter tuning to identify the baseline algorithm configuration. Secondly, we combined two different modifications to the benchmark technique, leading to consistent route length results. For future work, we suggest additional improvements, mainly considering the stochastic aspects of the method, and advanced techniques to choose the best hyperparameters, which may bring better results to the SOM algorithm.

\bibliographystyle{sbc}
\bibliography{sbc-template}

\end{document}